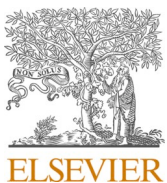
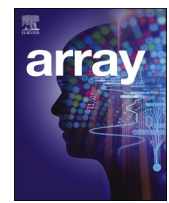

# CST-AFNet: A dual attention-based deep learning framework for intrusion detection in IoT networks

Waqas Ishtiaq [a], Ashrafun Zannat [a,b], A.H.M. Shahariar Parvez [c], Md. Alamgir Hossain [a,d,*], Muntasir Hasan Kanchan [a,d], Muhammad Masud Tarek [d]

[a] *Skill Morph Research Lab., Skill Morph, Dhaka, Bangladesh*
[b] *Department of Computer Science and Engineering, Bangladesh Army University of Science & Technology, Saidpur, Bangladesh*
[c] *Department of Software Engineering, Daffodil International University, Dhaka, Bangladesh*
[d] *Department of Computer Science and Engineering, State University of Bangladesh, Dhaka, Bangladesh*

## ARTICLE INFO

## ABSTRACT

The rapid expansion of the Internet of Things (IoT) has revolutionized modern industries by enabling smart automation and real-time connectivity. However, this evolution has also introduced complex cybersecurity challenges due to the heterogeneous, resource-constrained, and distributed nature of these environments. To address these challenges, this research presents CST-AFNet, a novel dual attention-based deep learning framework specifically designed for robust intrusion detection in IoT networks. The model integrates multi-scale Convolutional Neural Networks (CNNs) for spatial feature extraction, Bidirectional Gated Recurrent Units (BiGRUs) for capturing temporal dependencies, and a dual attention mechanism, channel and temporal attention to enhance focus on critical patterns in the data. The proposed method was trained and evaluated on the Edge-IIoTset dataset, a comprehensive and realistic benchmark containing over 2.2 million labeled instances spanning 15 attack types and benign traffic, collected from a seven-layer industrial testbed. Our proposed model achieves an outstanding accuracy with 15 attack types and benign traffic. CST-AFNet model achieves 99.97 % accuracy. Moreover, this model demonstrates an exceptional accuracy with macro-averaged precision, recall, and F1-score all above 99.3 %. Experimental results demonstrate that CST-AFNet achieves superior detection accuracy, significantly outperforming traditional deep learning models. The findings confirm that CST-AFNet is a powerful and scalable solution for real-time cyber threat detection in complex IoT/IIoT environments, paving the way for more secure, intelligent, and adaptive cyber-physical systems.

## 1. Introduction

Intrusion in IoT and IIoT refers to any unauthorized access, malicious activity, or attack that compromises the confidentiality, integrity, or availability of connected devices, systems, or data across industrial and consumer networks [1–3]. As IoT and IIoT infrastructures become increasingly integrated into critical domains such as manufacturing, energy, healthcare, and transportation, they present a vastly expanded attack surface due to their heterogeneous nature, resource-constrained devices, and reliance on lightweight protocols [4–6]. The problem we are solving in this research is the development of an intelligent and efficient intrusion detection mechanism capable of accurately identifying and classifying diverse cyber threats within these complex and dynamic IoT/IIoT environments.

The rapidly evolving threat landscape in IoT environments surpasses the capabilities of conventional security mechanisms, which were originally designed for static, homogenous, and resource-rich systems [7,8]. Unlike traditional networks, IoT/IIoT infrastructures operate under stringent constraints, including limited memory, processing power, and real-time responsiveness, making them highly vulnerable to novel and adaptive cyberattacks that can disrupt critical industrial operations, compromise safety, and cause significant economic and reputational damage. Furthermore, the increasing deployment of autonomous and mission-critical IoT systems demands proactive, intelligent, and context-aware security solutions [9–11].

Solving intrusion detection in IoT environments is inherently difficult due to their dynamic, distributed, and heterogeneous nature. These systems consist of diverse devices with varying computational capacities, communication protocols, and data formats, making unified security modeling a complex task. Moreover, the high volume and velocity of real-time data, coupled with the scarcity of labeled attack samples, especially for rare or stealthy intrusions, pose significant challenges for accurate threat detection. The presence of noise, imbalanced class distributions, and the necessity for low-latency processing further complicate the application of traditional or generalized deep learning techniques, demanding more specialized, adaptive, and resource-efficient solutions [12,13].

* Corresponding author. Skill Morph Research Lab, Skill Morph, Dhaka, Bangladesh.
  *E-mail addresses:* waqas.ishtiaq@gmail.com (W. Ishtiaq), spzannat@baust.edu.bd (A. Zannat), shahariar.swe@diu.edu.bd (A.H.M. Shahariar Parvez), alamgir.cse14.just@gmail.com (Md. Alamgir Hossain), muntasir@sub.edu.bd (M. Hasan Kanchan), tarek@sub.edu.bd (M. Masud Tarek).

Despite significant progress in the field of intrusion detection, current research exhibits notable limitations when applied to IoT environments. A prevailing gap lies in the over-reliance on obsolete or artificially synthesized datasets such as KDD99, NSL-KDD, or Bot-IoT which fail to reflect the real-time complexity, protocol diversity, and device heterogeneity characteristic of modern cyber-physical systems [14]. Consequently, many proposed models demonstrate high accuracy in controlled settings but struggle to generalize to real-world deployments. Moreover, a large body of existing work employs shallow machine learning techniques with manually engineered features, which are insufficient for capturing the high-dimensional and temporal dependencies inherent in network traffic data generated by IoT and IIoT devices [15]. Even contemporary deep learning approaches often overlook the importance of adaptive attention mechanisms, treating all input features and time steps with equal weight, thereby diminishing their ability to detect low-frequency, stealthy, or evolving threats. Additionally, most models are developed under the assumption of centralized data availability, ignoring the privacy constraints and computational limitations of edge-based or federated IIoT architectures, where data decentralization and bandwidth constraints are critical operational factors [16]. These limitations collectively indicate a significant research gap: the need for robust, scalable, and context-aware intrusion detection frameworks that are capable of learning from realistic, multi-modal data, extracting spatio-temporal patterns, and adapting intelligently to the dynamic threat landscape in resource-constrained, decentralized industrial environments.

To address the aforementioned gaps and limitations, we propose a novel deep learning framework named CST-AFNet (Convolutional Spatio-Temporal Attention Fusion Network), specifically designed for intrusion detection in complex IoT and IIoT environments. Our approach leverages a multi-scale convolutional architecture to automatically extract spatial features from diverse protocol layers and traffic patterns, effectively capturing localized attack signatures. To model the sequential nature of network behavior, we integrate a Bidirectional Gated Recurrent Unit (BiGRU) that captures both forward and backward temporal dependencies, enabling the detection of time-evolving and stealthy threats. Crucially, CST-AFNet introduces a dual attention mechanism comprising temporal attention, which allows the model to focus on the most informative time steps in a sequence, and channel attention, which highlights the most critical features contributing to an attack. This fusion of spatial, temporal, and attention-based modeling empowers the network to detect both high-frequency volumetric attacks and low-frequency, subtle intrusions with enhanced sensitivity. Key contributions of this research are listed below.

- We propose CST-AFNet, a novel deep learning architecture that fuses multi-scale CNN, BiGRU, and dual attention mechanisms for precise intrusion detection in IoT and IIoT networks.
- Our model integrates temporal and channel attention to selectively emphasize informative time steps and critical features, enhancing the detection of both frequent and rare cyberattacks.
- This research provides a scalable and edge-intelligent solution, paving the way for practical deployment of deep learning-based IDS in decentralized IoT environments.

Our proposed model captures spatial features at varying kernal size while BiGRU layers model temporal dependencies in network traffic data from both past and future contexts, allowing the detection of complex, time-dependent attack patterns. This selective emphasis enables the model to focus on the most informative temporal segments and critical characteristics that distinguish normal behavior from cyber-attacks. However, CST-AFNet effectively improves detection accuracy not only for common attack types but also for rare attack by traditional IDS methods. CST-AFNet develops an edge devices with limited resources, enabling real-time threat monitoring and response at the network's edge. This capability paves the way for practical, wide-scale adoption of deep learning-based IDS in diverse IoT infrastructures, enhancing cybersecurity resilience across interconnected smart systems.

The next section of this paper will present the Related Works, followed by the Proposed Methodology, then the Experimental Results and Analysis, and finally the Conclusion along with the References.

## 2. Related work

Intrusion Detection Systems (IDS) for IoT and IIoT networks have evolved rapidly to counter the increasing sophistication of cyberattacks targeting heterogeneous, decentralized, and resource-constrained environments. Traditional IDS approaches relied heavily on signature-based detection, which struggles to detect novel and zero-day attacks due to their rigid rule sets and static behavior models. As a response, machine learning (ML) and deep learning (DL) have emerged as powerful paradigms capable of learning complex patterns from network traffic data and enabling adaptive, data-driven intrusion detection [17].

Early works utilizing classical ML models such as Decision Trees, SVMs, and K-Nearest Neighbors demonstrated moderate success on legacy datasets like KDD99 and NSL-KDD, but failed to generalize to modern, real-time IIoT environments due to their limited scalability and dependence on manual feature engineering [18–20]. More recent efforts have applied DL architectures like CNNs, RNNs, and LSTMs to capture high-dimensional and temporal dependencies in traffic sequences, yielding improved detection accuracy [21–23]. However, these methods often overlook attention mechanisms and typically fail to perform well on minority-class attacks or under imbalanced data conditions [24,25].

Hybrid models have recently gained attention by combining CNNs with GRUs/LSTMs for joint spatial-temporal analysis [26]. Still, many such approaches lack an integrated attention mechanism to prioritize relevant features and time steps, limiting their interpretability and sensitivity to stealthy threats. Furthermore, most models are evaluated on synthetic or outdated datasets that do not reflect real-world IIoT infrastructure or traffic heterogeneity [27,28].

Another limitation in prior research is the lack of support for decentralized or privacy-aware learning. While some federated learning-based IDS frameworks have emerged [29], they often do not integrate deep spatio-temporal modeling or real-time adaptability, reducing their effectiveness in industrial applications. Additionally, explainability and resource efficiency critical for edge deployment are rarely addressed in existing solutions [30,31].

Recent developments have attempted to incorporate attention mechanisms (e.g., Graph Attention Networks and Temporal Attention) [32,33], optimize ensemble deep models, or utilize lightweight IDS in smart grids and edge computing contexts [34]. However, very few studies propose a unified architecture that integrates multi-scale CNNs, bidirectional temporal modeling, and dual attention mechanisms, especially one validated on a realistic and diverse dataset like Edge-IIoTset.

In [35], Data imbalance is a major concern for intrusion detection. However, they proposed a function to automatically capture negative negative cases over simple ones. This research demonstrates the convolutional neural network based intrusion detection system to enhance the detection of both frequent and rare cyberattacks. An ensemble learning methods is proposed to solve the data imbalance problem for intrusion detection dataset using Deep Neural Network (DNN) methods [36]. This study highlights the effectiveness of a deep learning-based intrusion detection approach that leverages dynamically scaled gradients through focal loss, enabling more efficient model training. The proposed method is evaluated across three datasets representing distinct IoT domains, demonstrating its adaptability and performance. Most of the works are done on centralized task offloading from mobile devices (MDs) to edge servers, limiting scalability and adaptability in dynamic environments. To address this, a decentralized task offloading framework formulated as a stochastic optimization problem under partial observability. Research in Ref. [37], the authors introduced a Mixed Multi-Agent Proxy Policy Optimization (mixed MAPPO) algorithm that enables efficient task distribution without inter-device coordination.

To fill these gaps, this research introduces CST-AFNet, a dual attention-based deep learning architecture that fuses multi-scale CNNs and BiGRU with temporal and channel attention mechanisms. The model is evaluated on Edge-IIoTset to demonstrate its robustness in detecting a wide range of attacks, including underrepresented classes, while maintaining scalability and interpretability for deployment in real-world IIoT scenarios.

## 3. Proposed Methodology

The methodology of this study involved three components: (1) proposed Multi stages CNN method, (2) added Bi-directional GRU, and (3) analysis the model with dual attention functions for the features extraction that can rank as the best classification for both frequent and rare cyberattacks.

Fig. 1 shows, we have been creating a multi stages CNN model that can concatenate with each other's. However, our proposed model also adds a type of recurrent neural network architecture is called Bi-directional GRU.

Then we create dual attention function while the model is extracting the most important features. Finally, we add Global Maxpooling1D layers, and a output layer is used for both binary and multiclass classification.

According to Table 1, the CST-AFNet model starts with three parallel Conv1D layers with different kernel sizes for multiscale feature extraction. The outputs are concatenated, normalized, and regularized with dropout. For capturing sequential patterns, we use BiGRU layer, followed by dual attention mechanisms to emphasize key temporal and channel-wise features. The final dense and softmax layers perform classification over 15 attack categories.

To capture the important features for detecting the different types of intrusion, we use a multi-scale convolutional approach using kernel sizes of 3, 5, and 7. Small kernels (e.g., $3 \times 3$) are used for detecting fine-grained local patterns such as subtle anomalies in packet-level traffic. Medium kernels ($5 \times 5$) expand the receptive field to identify mid-level features, while larger kernels ($7 \times 7$) enable the network to capture broader contextual information, such as aggregate behaviors and volumetric patterns characteristic of complex attacks like DDoS.

### 3.1. Dataset overview and preprocessing

The dataset used in this research is the Edge-IIoTset dataset [38], a comprehensive and realistic intrusion detection benchmark specifically designed for IoT security research. It comprises approximately 2,219, 201 network flow records, covering 15 attack types along with normal traffic, distributed across diverse communication protocols such as TCP, UDP, MQTT, HTTP, Modbus TCP, and ICMP. Each sample contains 63 features describing various packet-level and protocol-level behaviors.

The preprocessing involves the following steps.

i. Handling Missing Values: The dataset $D = \{x_i, y_i\}_{i=1}^{N}$, Where $x_i \in R^d$ and $y_i$ represent the features and the labels. For numerical columns $x^{(j)} \in R$, missing entries are imputed Using the median value: $x_i^{(j)} \leftarrow Median\ (x^{(j)})$. In addition to categorical columns, we set of categories as $x^{(k)} \in C$. where missing data elements are replaced with the mode: $x_i^{(k)} \leftarrow Mode\ (x^{(k)})$.

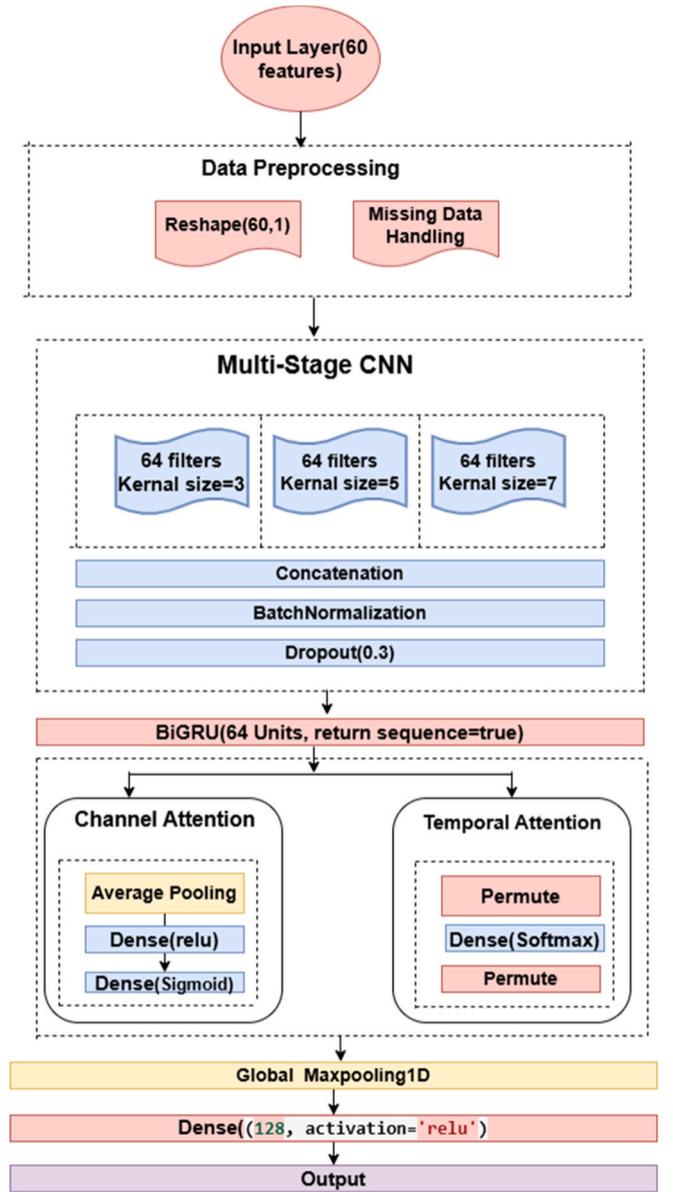

**Fig. 1.** Methodology for Intrusion Detection using a multi stages CNN model.

**Table 1**
Parameters of proposed CST-AFNet architecture.

| Component | Details/Values |
| --- | --- |
| **Input Shape** | (60) |
| **Reshape Layer** | Reshape to (60, 1) |
| **Conv1D - Kernel Size 3** | Filters: 64, Activation: ReLU, Padding: Same |
| **Conv1D - Kernel Size 5** | Filters: 64, Activation: ReLU, Padding: Same |
| **Conv1D - Kernel Size 7** | Filters: 64, Activation: ReLU, Padding: Same |
| **Concatenation** | Merge outputs from all three Conv1D layers |
| **Batch Normalization** | Applied after concatenation |
| **Dropout 1** | Rate: 0.3 |
| **BiGRU Layer** | Units: 64, Bidirectional, Return Sequences: True |
| **Temporal Attention** | Softmax attention across time dimension |
| **Channel Attention** | Dense layers: (64 → 8 → 64), Activation: ReLU + Sigmoid |
| **Global Max Pooling** | Aggregates features across time |
| **Dense Layer** | Units: 128, Activation: ReLU |
| **Dropout 2** | Rate: 0.4 |
| **Output Layer** | Units: 15, Activation: Softmax |

ii. Non-predictive or redundant features (e.g., Attack_label) are required to prevent bias or information leakage during training.
iii. At this stage, we normalize the data using StandardScaler, ensuring zero mean and unit variance: $x_i^{(j)} \leftarrow \frac{x_i^{(j)} - \mu_j}{\sigma_j}$, where, $\mu_j$ and $\sigma_j$ denote the mean and standard deviation of feature $j$ respectively.
iv. Finally, we convert the categorical fields into integer values using Label Encoding, and mapping each unique string to a unique integer: $Encode : C \rightarrow Z$

### 3.2. CST-AFNet architecture

The proposed CST-AFNet has designed to efficiently extract spatial, sequential, and attention-based representations from IoT/IIoT network traffic data.

The network consists of four main modules: Multi-Scale Convolutional Feature Extractor, Bidirectional GRU Temporal Modeler, Dual Attention Mechanism (Temporal and Channel Attention), Global Pooling and Final Classifier. The description of each component is given below.

#### 3.2.1. Input layer

Each input instance $x \in R^d$ (Where, d = No. of features after pre-processing) is reshaped into a 2D tensor for 1D convolution processing: $x \in R^{d \times 1}$.

This transformation is identical for both binary and multiclass classification tasks, with the only variation occurring in the output label space, i.e., $y \in \{0,1\}$ for binary and $y \in \{0,1,\ldots,C-1\}$ for multiclass, where $C$ is the number of classes.

In this proposed system, we have extract 60 different features from our dataset, those features help to the detection of both frequent and rare cyberattacks.

#### 3.2.2. Multi-scale convolutional block

To capture features at different spatial resolutions, the CST-AFNet model has three parallel 1D Convolutional Neural Networks (CNNs). We are applied those layers with different sizes kernel $k = \{3,5,7\}$. Each convolution operation for a given kernel size $k$ is defined as:

$$h_k = ReLU\ (Conv1D_k(x))$$

Where, $Conv1D_k$ is a $1D$ convolution with filter size $k$. ReLU is the activation function. For each layer, the output have 64 features map.

Then the model performs a concatenated operation with the output of three convolutions layers. The function for the concatenation is given bellow:

$$h_{concat} = Concat\ (h_3, h_5, h_7)$$

Batch Normalization layer and Dropout (rate = 0.3) are then applied to $h_{concat}$. After that, it is stabilized for the training and prevent overfitting.

#### 3.2.3. Bidirectional GRU Temporal Modeler

The output $h_{concat}$ is fed into a Bidirectional Gated Recurrent Unit (BiGRU) to model sequential dependencies from both forward and backward directions: $h_{rnn} = BiGRU(h_{concat})$.

Where, each GRU unit processes the sequence as:

$$z_t = \sigma(W_z \cdot [h_{t-1}, x_t]), r_t = \sigma(W_r \cdot [h_{t-1}, x_t]),$$
$$\widetilde{h}_t = \tanh\ (W_z \cdot [r_t \odot h_{t-1}, x_t]), h_t = (1 - z_t \odot h_{t-1} + z_t \odot \widetilde{h}_t)$$

Where, $z_t$ is the update gate, $r_t$ is the reset gate, $h_t$ is the hidden state at time $t$, $\odot$ denotes element-wise multiplication, $\sigma$ denotes the sigmoid activation.

#### 3.2.4. Dual attention mechanism

After temporal modeling, dual attention mechanisms are applied to focus on the most relevant features and time steps.

(a) Temporal Attention

Temporal attention is added to our proposed CST-AFNet model to enhance the model's ability that can focus on the most relevant time steps in a sequential input. At first the model permutes the input sequence to switch the time and the feature dimension, then it allows a dense layer with a softmax activation function to learn weights across the temporal axis. When the attention scores are computed, the tensor is permuted back to its original shape. Finally, the original input and the attention weights are element-wise multiplied with each other's. This result is effectively highlighting the significant temporal features while suppressing less relevant ones. This mechanism allows the model to dynamically prioritize important temporal patterns, which is especially useful in time-series or sequential data analysis.

Focuses on important time steps in the sequential output $h_{rnn}$. Formally:

$$A_T = softmax(W_T \cdot h_{rnn}), h_{temp} = h_{rnn} \odot A_T$$

Where, $A_T$ is the attention score matrix for time steps, $h_{temp}$ is the attended temporal representation.

(b) Channel Attention

In this research, we used a channel attention feature channel to emphasize the most relevant feature maps and suppresses less relevant ones. It begins with global average pooling across a Lambd layer, which can compress the input to create a summary of each feature channel. This function creates through to pass two fully connected (Dense) layers: the first reduces dimensionality (using a reduction_ratio) and applies a ReLU activation to capture non-linear interactions, while the second restores the original channel dimension and uses a sigmoid activation to generate channel-wise attention weights between 0 and 1. Finally, these weights are required to multiple element-wise with original input. Focuses on important channels/features in $h_{temp}$. It involves:

1. Global Average Pooling along the temporal axis: $s = \frac{1}{T} \sum_{t=1}^{T} h_{temp}(t)$
2. Passing through two dense layers (bottleneck reduction by ratio ($r$): $s' = \sigma(W_2 \cdot ReLU(W_1 \cdot s))$
3. Rescaling: $h_{chan} = h_{temp} \odot s'$

Where, $W_1$ and $W_2$ are trainable parameters, $\sigma$ is the sigmoid activation function, $T$ is the number of time steps.

#### 3.2.5. Global Pooling and Final Classification

The final attended feature map $h_{chan}$ is passed through a GlobalMaxPooling1D layer to obtain a fixed-size feature vector: $h_{pool} = GlobalMaxPool(h_{chan})$. This pooled representation is then processed through a Dense layer with 128 neurons and ReLU activation, followed by a Dropout layer with a rate of 0.4 to prevent overfitting. Finally, an output layer is applied to produce the class probabilities or decision score. For binary classification, the output layer uses a single neuron with a sigmoid activation: $\widehat{y} = \sigma(W_{out} \cdot h_{pool} + b_{out})$.

For multiclass classification (e.g., 15 classes in Edge-IIoTset), the output layer contains $C$ neurons (where $C$ is the number of classes) with a softmax activation: $\widehat{y}_j = \frac{e^{(W_{out,j} \cdot h_{pool} + b_{out,j})}}{\sum_{k=1}^{C} e^{(W_{out,k} \cdot h_{pool} + b_{out,k})}}, j = 1, \ldots, C.$

Where, $W_{out}$ and $b_{out}$ are weights and bias of the final layer and $\widehat{y}$ or $\widehat{y}_j$ represents the predicted probability for binary or each class, respectively. This final classification layer enables the model to adapt flexibly to either binary decision-making (attack vs. normal) or fine-grained multiclass threat identification.

### 3.3. Model training strategy

The training strategy of the proposed CST-AFNet model is carefully designed to ensure robust convergence, mitigate overfitting, and maximize generalization performance on the dataset. To ensure robust, we proposed an adaptive learning strategy with different hyperparameters. We have selected a callbacks function to optimize performance and prevent overfitting. Moreover, the Adam optimizer provides efficient gradient-based updates with an initial learning rate of 0.001. Furthermore, this learning rate helps to prevent our loss values. Sparse Categorical Crossentropy is used for multi-class classification with integer-encoded labels. We use a large batch size to stabilize gradient updates and train for a maximum of 10 epochs, with early stopping monitoring the validation loss (patience = 5) and restoring the best weights. The trained parameters with their values are presented Table 2.

Fig. 2 illustrates the CST-AFNet Model training and compiled steps. The dataset is divided into train and validation folder with 80 % and 20 % ratio. The 80:20 train-test split was chosen based on standard practices in machine learning research, as it offers a good balance between sufficient training data for deep model convergence and enough test data to evaluate generalization. This ratio has also been used in prior IoT intrusion detection studies [17,36] ensuring comparability and reliability of performance metrics.

#### 3.3.1. Loss function

For binary classification problem (Normal vs. Attack), the Binary Cross-Entropy (BCE) loss function is employed.

The loss for a single instance is defined as:

$$\mathscr{L}_{BCE}(y, \widehat{y}) = -(y\log(\widehat{y}) + (1-y)\log(1-\widehat{y}))$$

Where, $y \in \{0,1\}$ is the true binary label, $\widehat{y} \in \{0,1\}$ is the predicted probability output from the sigmoid activation. The overall loss minimized during training is the empirical average: $\mathscr{L}_{total} = \frac{1}{N}\sum_{i=1}^{N}\mathscr{L}_{BCE}(y_i, \widehat{y}_i)$.

For multiclass classification (e.g., 15 attack categories), the Sparse Categorical Cross-Entropy (SCCE) is used. The loss for an instance is defined as: $\mathscr{L}_{SCCE}(y, \widehat{y}) = -\log(\widehat{y}_y)$, where, $y \in \{0,1,\ldots,C-1\}$ is the true class label (as an integer), $\widehat{y} \in R^C$ is the output softmax probability vector, $\widehat{y}_y$ is the predicted probability corresponding to the true class index. The total loss over $N$ samples is: $\mathscr{L}_{total} = \frac{1}{N}\sum_{i=1}^{N}\mathscr{L}_{SCCE}(y_i, \widehat{y}_i)$. This adaptive loss selection allows CST-AFNet to function seamlessly for both binary and multiclass intrusion detection tasks using the appropriate probabilistic output layer and loss formulation.

**Table 2**
CST-AFNet Model trained Parameters.

| Parameters | Values |
|---|---|
| Optimizer | Adam |
| Learning Rate | 0.001 |
| Loss Function | Sparse Categorical Crossentropy |
| Metric | Accuracy |
| Batch Size | 1024 |
| Epochs | 10 |
| Validation Split | 0.2 |
| Early Stopping | Monitor: val_loss, Patience: 5, Restore Best Weights |
| Model Checkpoint | Save Best Only: best_cst_afnet_model.h5 |
| Train/Test Split Ratio | 80 % training, 20 % testing (with stratification) |
| Label Encoding | LabelEncoder used for class labels |

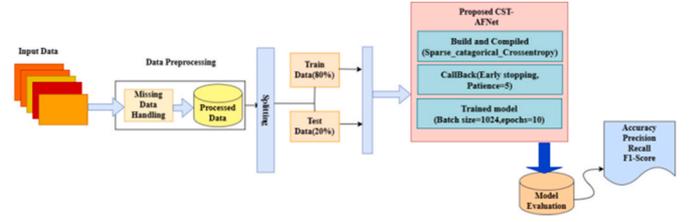

**Fig. 2.** Training and model evaluation steps using our proposed CST-AFNet model.

#### 3.3.2. Optimizer

The model parameters are optimized using the Adam Optimizer, a widely adopted method in deep learning due to its adaptive learning rate and momentum capabilities. Adam updates parameters $\theta$ at iteration $t$ according to:

$$m_t = \beta_1 m_{t-1} + (1-\beta_1)\nabla_\theta \mathscr{L}_t, v_t = \beta_2 v_{t-1} + (1-\beta_2)(\nabla_\theta \mathscr{L}_t)^2,$$

$$\widehat{m}_t = \frac{m_t}{1-\beta_1^t}, \widehat{v}_t = \frac{v_t}{1-\beta_2^t}, \theta_t = \theta_{t-1} - \eta\frac{\widehat{m}_t}{\sqrt{\widehat{v}_t}+\epsilon}$$

Where, $\eta = 10^{-3}$ (learning rate), $\beta_1 = 0.9, \beta_2 = 0.999, \epsilon = 10^{-8}$ (numerical stability).

#### 3.3.3. Training Configuration

A batch size of 1024 is used, balancing memory efficiency with gradient stability. The maximum number of epochs is set to 10, although training is dynamically halted earlier if no improvement is observed. 20 % of the training set is reserved for validation during each epoch.

Two important callbacks are integrated during training. Early Stopping Monitors validation loss. If validation loss does not improve for 5 consecutive epochs, training is halted. The best weights (based on lowest validation loss) are restored automatically. Model Checkpoint saves the model weights whenever the validation loss improves. Ensures that the best-performing model during training is preserved.

### 3.4. Evaluation metrics

To evaluate the performance of the proposed CST-AFNet model, we utilize the classification report and confusion matrix two widely adopted metrics in intrusion detection tasks. The classification report provides key indicators including precision, recall, and F1-score for each class, offering insights into the model's effectiveness in correctly identifying both normal and attack instances, particularly in the presence of class imbalance. The confusion matrix further complements this analysis by detailing the true positives, true negatives, false positives, and false negatives, enabling a granular understanding of misclassification patterns.

## 4. Results & analysis

All experiments were conducted in Google Colab using Python 3 and the TensorFlow/Keras deep learning framework. The proposed CST-AFNet model was implemented with core layers such as Conv1D, GRU, Dense, Dropout, BatchNormalization, and custom attention blocks (channel_attention, temporal_attention). Data preprocessing utilized pandas, StandardScaler, and LabelEncoder from sklearn. The model was trained using the Adam optimizer (learning rate = 0.001), with a batch size of 1024, 10 epochs, and EarlyStopping and ModelCheckpoint callbacks for regularization. Evaluation was performed using classification_report and confusion_matrix from sklearn.metrics.

Using precision in Fig. 4, recall in Fig. 3, and F1-score in Fig. 5 as metrics, Table 3 represents the evaluation of the model's effectiveness across various cyberattack categories. The results indicate that our proposed model performs outstanding, achieving a great score in precision, recall, and F1-score for several critical classes such as DDoS_HTTP, DDoS_TCP, DDoS_UDP, MITM, Normal, and Port_Scanning, suggesting that the model accurately identifies these threats without any misclassifications.

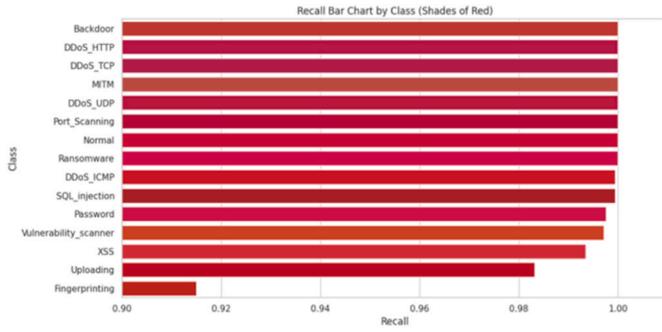

**Fig. 3.** Recall for each class in the multi-class classification.

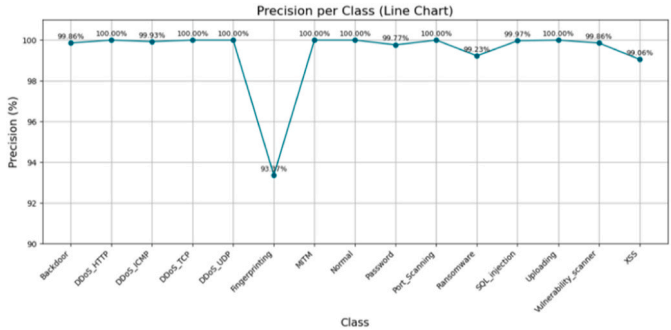

**Fig. 4.** Precision for each class in the multi-class intrusion detection.

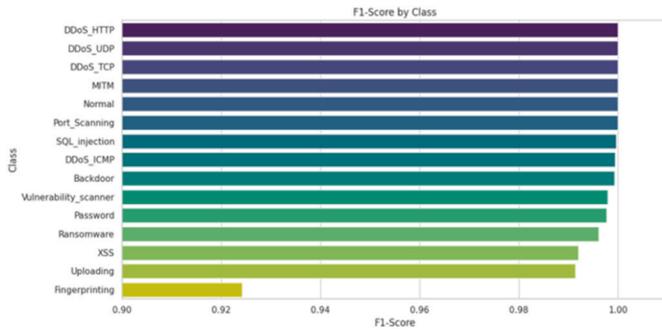

**Fig. 5.** F1-score for each class in the multi-class intrusion detection.

**Table 3**
Classification report for multiclass classification.

| Class/Metric | Precision | Recall | F1-Score |
|---|---|---|---|
| Backdoor | 0.9986 | 1.0000 | 0.9993 |
| DDoS_HTTP | 1.0000 | 1.0000 | 1.0000 |
| DDoS_ICMP | 0.9993 | 0.9994 | 0.9994 |
| DDoS_TCP | 1.0000 | 1.0000 | 1.0000 |
| DDoS_UDP | 1.0000 | 1.0000 | 1.0000 |
| Fingerprinting | 0.9337 | 0.9150 | 0.9242 |
| MITM | 1.0000 | 1.0000 | 1.0000 |
| Normal | 1.0000 | 1.0000 | 1.0000 |
| Password | 0.9977 | 0.9976 | 0.9977 |
| Port_Scanning | 1.0000 | 1.0000 | 1.0000 |
| Ransomware | 0.9923 | 1.0000 | 0.9961 |
| SQL_injection | 0.9997 | 0.9994 | 0.9996 |
| Uploading | 1.0000 | 0.9831 | 0.9914 |
| Vulnerability_scanner | 0.9986 | 0.9971 | 0.9979 |
| XSS | 0.9906 | 0.9934 | 0.9920 |
| **Accuracy** | **0.9997** | | |
| **Macro Average** | **0.9940** | **0.9933** | **0.9937** |
| **Weighted Average** | **0.9997** | **0.9997** | **0.9997** |

Here, fingerprinting showed relatively lower F1-scores compared to others, it indicates that subtle and stealthy attacks may still pose a challenge.

Moreover, we address a great model complexity and computational overhead, due to attention mechanisms and deep layers might limit its deployment in resource-constrained edge devices typical in IIoT settings.

Backdoor, DDoS_ICMP, SQL_injection, Password, Vulnerability_scanner, and Ransomware are also performed an accurate performance with very high metric values, and those attack reflect the model's robustness in recognizing their impact. However, a relatively lower performance is observed in the Fingerprinting class, where precision (0.9337), recall (0.9150), and F1-score (0.9242) are slightly reduced, indicating some misclassifications. The classification performance of the proposed CST-AFNet model on multiclass intrusion detection. The model achieves an exceptional overall accuracy of 99.97 %, with macro-averaged precision, recall, and F1-score all above 99.3 %. The macro and weighted average values are presented in Fig. 7. It effectively detects all attack types, including rare ones like MITM and Fingerprinting, confirming its robustness and precision across diverse and imbalanced IIoT traffic scenarios.

Fig. 6 illustrates the confusion matrix for multiclass classification using the proposed CST-AFNet model. The results show near-perfect classification across all classes, including highly imbalanced ones like MITM, Fingerprinting, and XSS, demonstrating the model's strong generalization and precise discrimination of complex attack patterns in IIoT environments.

Fig. 8 shows, the training and validation accuracy of the CST-AFNet model shows using a plot against over 10 epochs. This model demonstrates a rapid increase in both training and validation accuracy within the first few epochs, stabilizing above 99 % accuracy. This indicates that the model is learning effectively and generalizing well.

At the early stage, the validation accuracy remains closely aligned with training accuracy, we see a minor fluctuation but it does not mean an overfitting issue. The loss plot (right) shows a steep decline in both training and validation loss within the first two epochs, followed by a gradual convergence to near-zero values. A small spike in validation loss is noticed around epochs 6–7, but the loss quickly recovers and continues to decrease. Finally, the CST-AFNet model achieves fast convergence and maintains strong generalization performance throughout the training process.

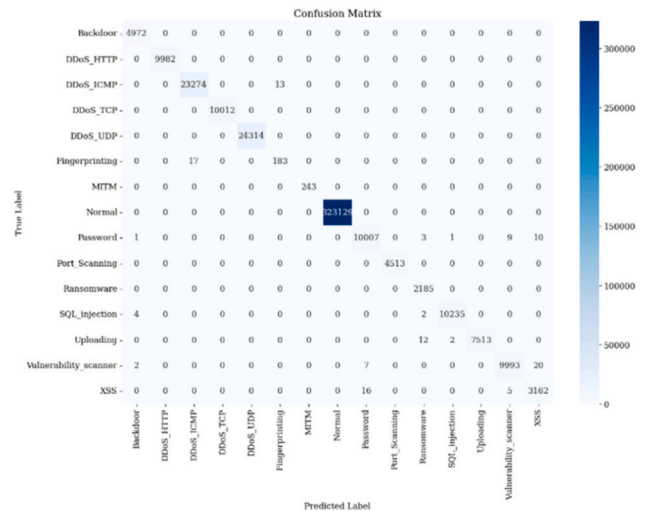

**Fig. 6.** Confusion matrix for multiclass classification.

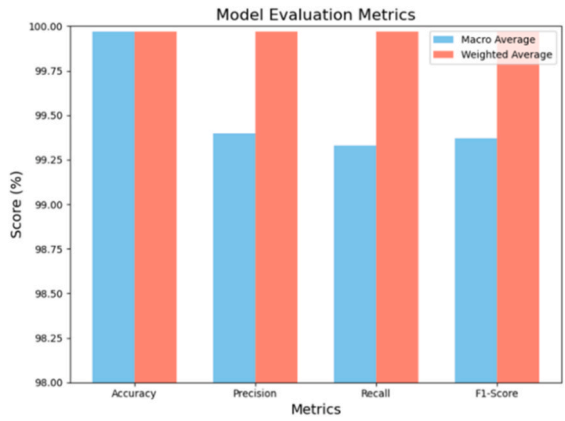

**Fig. 7.** Marco Average and weighted Average values for Multi-class Classification.

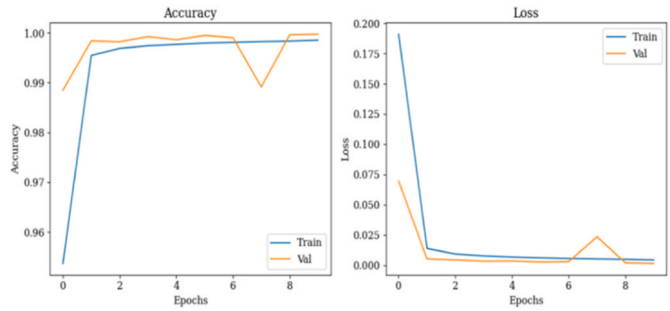

**Fig. 8.** Accuracy and Loss curve against number of epochs for multi class intrusion detection.

**Table 4**
Classification report for binary (anomaly) class classification.

| Class/Metric | Precision | Recall | F1-Score |
| --- | --- | --- | --- |
| Attack | 1.0000 | 1.0000 | 1.0000 |
| Normal | 1.0000 | 1.0000 | 1.0000 |
| **Accuracy** | **1.0000** | | |
| **Macro Average** | **1.0000** | **1.0000** | **1.0000** |
| **Weighted Average** | **1.0000** | **1.0000** | **1.0000** |

Table 4 shows perfect performance of the proposed model in binary classification, achieving 100 % precision, recall, and F1-score for both attack and normal classes, confirming its exceptional accuracy in anomaly detection.

Fig. 9 displays the confusion matrix for binary classification, where the CST-AFNet model perfectly distinguishes between normal and attack traffic with zero misclassifications, confirming its exceptional reliability for anomaly detection.

The training history plots in Fig. 10 illustrate the performance of a binary classification model is shown over 7 epochs. However, both training and validation accuracy rapidly reach 100 %. In this research, the validation accuracy remains relentlessly flat at 1.0 across all epochs.

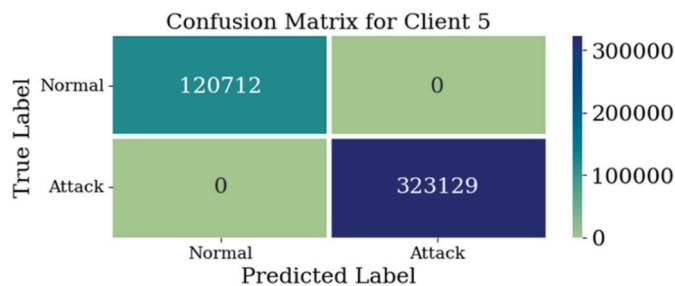

**Fig. 9.** Confusion matrix for binary (anomaly) class classification.

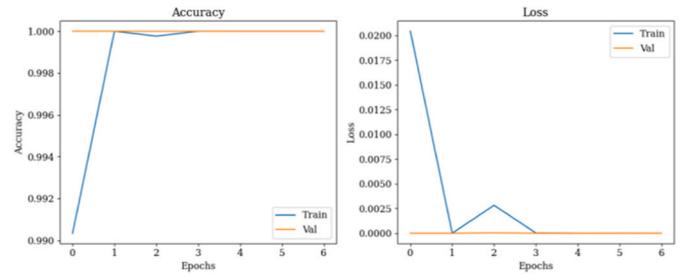

**Fig. 10.** Accuracy and Loss curve against number of epochs for binary class intrusion detection.

**Table 5**
The state-of-the-art analysis for IDS and CST-AFNet model.

| Reference | Classifier | Dataset | Accuracy |
| --- | --- | --- | --- |
| [35] | FNN-Focal | Bot-IoT | 0.9155 |
| [35] | CNN-Focal | Bot-IoT | 0.8677 |
| [39] | CNN + LSTM | UNSW-NB15 | 0.929 |
| [40] | RaNN | ToN_IoT | 0.9905 |
| [41] | DNN | – | 0.8400 |
| [41] | GAN-DNN | UNSW-NB15 | 0.9100 |
| [36] | DNN | NSL-KDD | 0.9890 |
| [36] | DNN | UNSW-NB15 | 0.9670 |
| [36] | DNN | CIC-IDS-2017 | 0.9874 |
| [36] | DNN | BOT-IOT | 0.9899 |
| [36] | RNN | – | 0.9958 |
| [36] | CNN-BiLSTM | – | 0.9976 |
| [42] | LSTM | – | 0.9973 |
| [42] | CNN | CICIDS2017 | 0.9882 |
| [43] | DNN | DS2OS Traffic | 0.9490 |
| [43] | FFNN | – | 0.9867 |
| [43] | LSTM | NSL-KDD | 0.9644 |

Our proposed model shows almost zero validation loss throughout the training process. Table 5 shows our proposed model performs outstanding than the existing methods for intrusion detection.

## 5. Conclusion & future work

This research introduced CST-AFNet, a novel dual attention-based deep learning model for intrusion detection in IoT and IIoT environments. The study explored both binary and multiclass classification on the Edge-IIoTset dataset, leveraging multi-scale CNNs, BiGRU, and temporal-channel attention mechanisms to extract meaningful spatio-temporal patterns from traffic data. The model demonstrated outstanding performance, achieving up to 100 % accuracy in binary classification and 99.97 % in multiclass classification, including excellent results on minority classes. These findings highlight the effectiveness and robustness of CST-AFNet in detecting diverse and complex attacks, making it a strong candidate for real-world IIoT security deployment. Future work will explore the deployment of CST-AFNet in federated and edge computing environments to enhance privacy and scalability. Additionally, integrating online learning and explainable AI techniques will improve adaptability to evolving threats and interpretability for security analysts. Finally, the model's robustness against adversarial attacks or data poisoning was not assessed, which is critical for security-critical environments like industrial systems. Future work should address these concerns by incorporating continual learning, adversarial robustness, and real-time deployment evaluation.

**CRediT authorship contribution statement**

**Waqas Ishtiaq:** Writing – original draft, Methodology, Conceptualization, Data curation. **Ashrafun Zannat:** Writing – original draft, Visualization, Validation, Methodology, Investigation, Formal analysis, Data curation, Conceptualization. **A.H.M. Shahariar Parvez:** Formal analysis, Supervision, Visualization. **Md. Alamgir Hossain:** Writing – original draft, Methodology, Investigation, Formal analysis, Data curation, Conceptualization. **Muntasir Hasan Kanchan:** Writing – review & editing, Writing – original draft, Visualization, Validation, Supervision, Methodology, Investigation, Data curation. **Muhammad Masud Tarek:** Writing – review & editing, Visualization, Validation, Supervision, Resources, Methodology.

**Funding**

**Declaration of competing interest**

The authors declare that they have no known competing financial interests or personal relationships that could have appeared to influence the work reported in this paper.

**Data availability**

Data will be made available on request.